\documentclass[conference]{IEEEtran}
\IEEEoverridecommandlockouts
\usepackage{times}
\usepackage[numbers]{natbib}
\usepackage{multicol}
\usepackage[bookmarks=true]{hyperref}
\usepackage{amsmath,amssymb,amsfonts}
\usepackage{graphicx}
\usepackage{textcomp}
\usepackage{xcolor}
\usepackage{nicefrac}
\usepackage{cleveref}
\usepackage{comment}
\usepackage{url}
\usepackage[subtle, tracking=normal, floats=normal]{savetrees}
\usepackage{algorithm}
\usepackage{algpseudocode}

\begin{document}

\title{Graph-Based Design of Soft Grippers with \\ Multi-Objective Quality-Diversity Optimisation

}
\author{\IEEEauthorblockN{Andr\'e Farinha\IEEEauthorrefmark{1}, Ge Shi\IEEEauthorrefmark{1}, Harry Bowman\IEEEauthorrefmark{1}, Brendan Tidd\IEEEauthorrefmark{1}, David Howard\IEEEauthorrefmark{1} and Josh Pinskier\IEEEauthorrefmark{1}}
\IEEEauthorblockA{\IEEEauthorrefmark{1} CSIRO Robotics, Australia | Contact: andre.farinha@csiro.au}  
}

\maketitle

\begin{abstract}
Effective manipulation across diverse objects is critical for applications ranging from agricultural harvesting to laboratory and domestic automation. While the inherent compliance of soft robotics is well suited to this challenge, designing grippers that generalize across tasks remains difficult due to the vast design space of continuum mechanics and the risk of overfitting to specific scenarios.

We propose a graph-based design space for representing soft structures and mechanisms, coupled with a multi-objective, diversity-driven genetic optimization framework that explicitly promotes solution variety throughout the design process. Using multiple grasping scenarios during optimization, we study how task diversity influences the emergence of generalization to unseen objects and contact conditions.

Our results show that optimization over a sufficiently diverse set of grasping cases leads to designs with emergent generalization, exhibiting improved robustness compared to task-specific solutions on novel scenarios. These findings suggest that diversity-driven optimization offers a principled pathway toward general-purpose soft grippers, aligned with the adaptable nature of soft robotics.

\end{abstract}

\section{Introduction}
\label{sec:intro}

Effective manipulation across variable objects is essential across numerous domains, from harvesting of diverse produce in agriculture, to managing equipment in laboratory automation. A world of ubiquitous robotics requires systems capable of interacting robustly with broad distributions of objects and equipment, rather than narrowly defined instances.

While recent advances in vision-language-action models \cite{pi0} demonstrate one path toward general manipulation, soft grippers offer a complementary approach that leverages mechanical intelligence rather than (or in addition to) computational intelligence \cite{whitesides2018soft}. Unlike rigid grippers requiring precise positioning, soft grippers accommodate geometric variation through material deformation and maintain stable grasps despite positioning errors. When properly designed, such grippers should naturally generalize, simplifying control and reducing reliance on task-specific learning.

However, realizing this promise hinges critically on the design of the soft structure itself. The design space of soft robotic systems is effectively unbounded, making computational optimal design essential \cite{Pinskier2022}. At the same time, the compliance and adaptability that enable robust grasping require expensive non-linear simulations to capture large deformations, material non-linearities, instabilities, and contact interactions \cite{whitesides2018soft}. This exposes a fundamental trade-off: restricting design spaces to remain compatible with high-fidelity simulation limits expressiveness, while exploring richer spaces using simplified physics neglects the phenomena that ultimately govern soft gripper performance \cite{Pinskier2022}.

In response, recent work has expanded computational design frameworks toward more expressive representations \cite{navarro2023open}. Topology optimization approaches \cite{liu2024topology, pinskier2024diversity}, for example, impose minimal structural assumptions but remain limited in their treatment of material boundaries, strong non-linearities, and contact \cite{bluhm2021internal}. Graph-based design spaces offer a promising middle ground, combining high expressiveness with structures compatible with available continuum simulation frameworks. In robotics, graph representations have enabled efficient evolutionary search over complex morphologies \cite{wang2019neural}, while spatial graphs have a long history in structural optimization \cite{hagishita2009topology} and renewed interest through graph neural network approaches \cite{kupwiwat2024multi}.

Despite these advances, incorporating rich physical behaviours within expansive design spaces remains challenging. High-fidelity FEM-based approaches often force optimization into narrowly parametrized families such as fin-ray-inspired structures \cite{xing_finray}. Moreover, even when high-fidelity simulation becomes tractable, coupling these models with conventional single-objective optimization frequently yields narrow and brittle solutions \cite{koos2012transferability, scheper2017abstraction}. This bespokeness is not solely a consequence of imperfect physics models, but also of the optimization process itself, which rewards exploitation over coverage and suppresses alternative high-performing designs.

To counter this effect, diversity-based optimization methods have emerged as essential tools. Quality-diversity algorithms such as MAP-Elites \cite{mouret2015illuminating} maintain an archive of high-performing solutions distributed across behavioural space, preventing premature convergence to narrow optima, while multi-objective evolutionary algorithms \cite{deb2002fast} preserve diverse Pareto-optimal trade-offs. Together, these approaches have demonstrated success in discovering robust and diverse robotic behaviours, including recent applications to soft robotic grasping that uncover diverse strategies without explicit task specification \cite{ pinskier2024diversity}, suggesting a pathway toward general-purpose soft manipulation that leverages complex non-linear phenomena without task-specific overfitting.

\begin{figure*}[]
    \centering
    \includegraphics[width=1\textwidth]{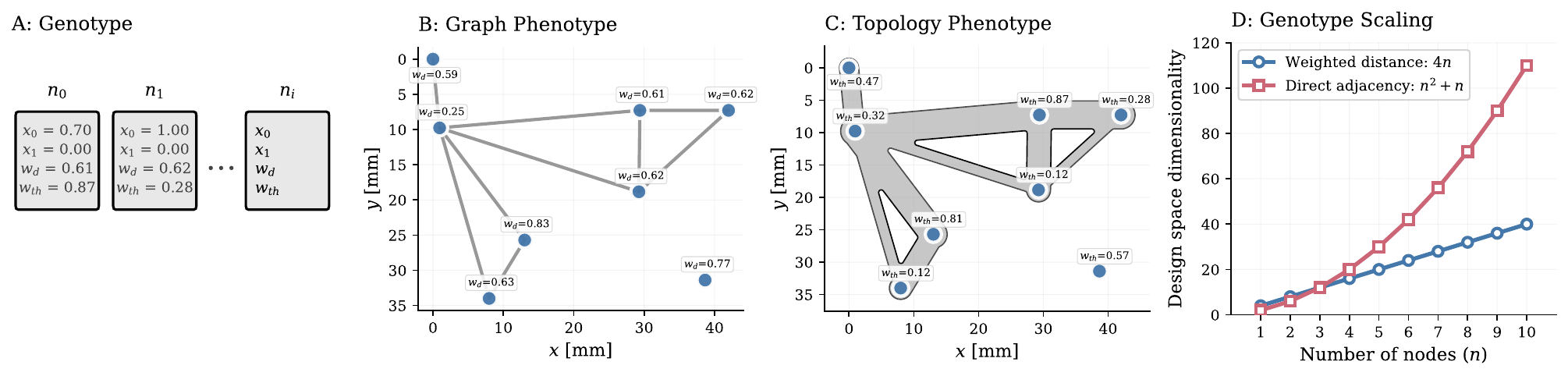}
    \vspace{-22pt}
    \caption{A. Genotype, B. Graph phenotype and C. Topology phenotype of a randomly generated graph. D. scaling of the thresholded weighted distance approach used in this work, and a direct adjacency encoding. }
    \label{fig:fig1}
    \vspace{-16pt}
\end{figure*}

In this work, we present a framework for near-freeform design of soft grippers that combines (1) expressive design representations, (2) high-fidelity physics, and (3) diversity-driven optimization. We propose a graph-based design space for planar soft structures and compliant mechanisms compatible with finite element simulation, and introduce a multi-objective quality-diversity algorithm based on Dominated Novelty Search \cite{bahlous2025dominated}. We validate the approach on benchmark structural optimization problems and soft gripper design tasks, and investigate whether optimization across diverse grasping scenarios induces emergent generalization to novel objects and contact conditions.

\section{Design Representation}
\label{sec:space}
As discussed above, graph-based representations offer an expressive framework for encoding complex morphologies while remaining amenable to automated search. In this work, we aim at achieving a similar level of expressiveness as found in voxel-based methods, but with increased tractability. While it might be challenging to contain as much information in a graph as one would in a voxel space, structures obtained from topology optimization, even for complex compliance-based problems \cite{koppen2022simple}, can often be approximated by graph structures. In fact, a graph representation where edges encode curved beams of variable thickness, has been previously used to optimise compliant mechanisms \cite{sauter2008complex}.  In this work, we simplify edges to straight beams of constant thickness, and instead achieve greater complexity through a graph representation that is compatible with a larger number of nodes. To better illustrate the design representation employed in this work, we introduce the following terminology: 

\textbf{Genotype.} The genotype is the internal representation of a design that the optimisation algorithm operates on. For this reason, it must be information-dense and searchable by optimisation algorithms. As shown in \cref{fig:fig1}.A the genotype used in this work is composed of a set of $n$ nodes as $[\boldsymbol{x}, w_d, w_{th}]_{n}$, where $\boldsymbol{x}$ are the Cartesian coordinates in 2D space, $w_d$ a distance multiplier, and $w_{th}$ an edge width multiplier. The genotype is thus of size $4n$ and bound in $[0,1]$. 

\textbf{Graph Phenotype.} The graph phenotype (\cref{fig:fig1}.B) is an undirected spatial graph $G = (V, E)$, where \textit{V} is a set of nodes and \textit{E} a set of edges. We choose to not directly encode the graph edges in the genotype to reduce genotype dimensionality. Instead, the set of active graph edges is inferred from the node coordinates and distance weights. This is achieved by thresholding the graph distance matrix $L_{ij}$ by enforcing $\{L_{ij}:j \ne i\} \leqslant d_{max} : d_{max} \in [0,1]$, where $d_{max} \in [0,1]$ is a hyperparameter representing the furthest possible link as a fraction of the domain diagonal. As a threshold purely based on Euclidean distance would severely restrict the set of possible designs, we use a distance multiplier to distort the distance matrix $\{L_{ij}:j \ne i\} = \| \boldsymbol{x_i} - \boldsymbol{x_j} \| + w_i w_j$. As shown in \cref{fig:fig1}.D this process allows for the genotype to be considerably denser than the alternative of directly encoding the graph adjacency matrix. Although this advantage diminishes for extremely sparse graphs, the representation remains efficient for this work. Furthermore, while this representation spans only a subset of the $2^{\frac{n}{2}(n-1)}$ graph topologies with $n$ nodes, our empirical results indicate that this subset is sufficiently expressive for the class of problems considered in this work. 

\textbf{Topology Phenotype.} The topology phenotype \cref{fig:fig1}.C is a planar geometric object resulting from the union of the graph edges in 2D space. As demonstrated in \cref{fig:fig1}.D, each graph edge fills the domain space with material in the shape of a straight line of thickness $w_{th} \cdot th_{max}$, where $th_{max}$ is the maximum allowable edge thickness. 

\begin{figure*}[]
    \centering
    \includegraphics[width=1\textwidth]{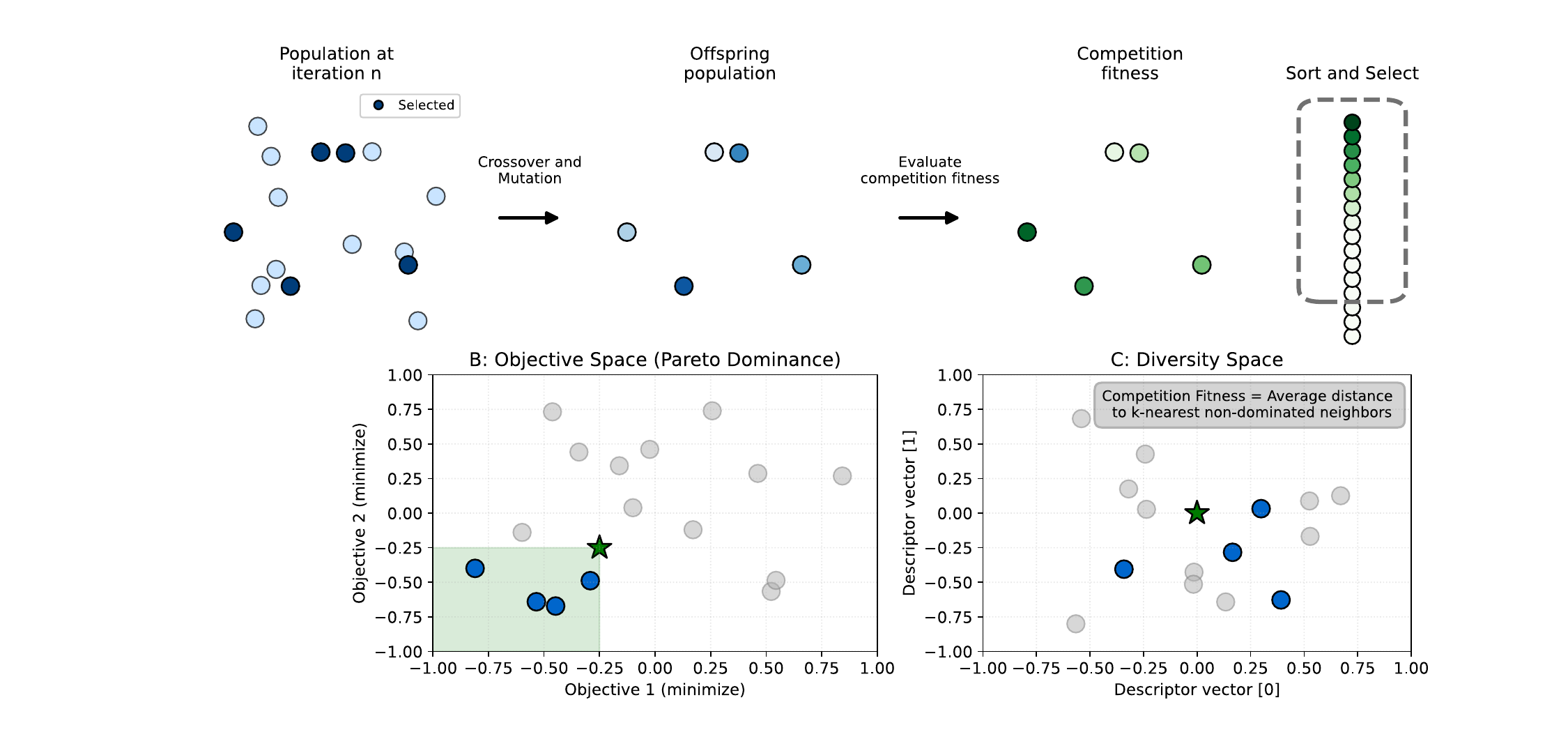}
    \vspace{-22pt}
    \caption{Overview of Pareto Dominated Novelty Search. Algorithm workflow: Selected individuals (dark blue) are reproduced and evaluated for fitness (blue gradient) and competition fitness (green gradient). B: Competition fitness calculation for a reference individual. Non-dominated individuals (blue) Pareto-dominate the reference in objective space. C: Competition fitness equals average distance to k-nearest non-dominated neighbours in diversity space. }
    \vspace{-12pt}
    \label{fig:fig2}
\end{figure*}

It should be noted that, although this work considers only planar flexible mechanisms represented by the topology phenotype, the proposed genotype and graph-based phenotype are sufficiently general to be applied to a broader range of robotics problems. Consequently, the optimisation approach described below can be adapted to a variety of problem settings with minimal modification.

\section{Multi-Objective Dominated Novelty Search}

The design space above, though expressive, is extremely vast, unbounded in the number of nodes, and unsurprisingly deceptive for standard optimisation algorithms to explore. As discussed in \cref{sec:intro}, multi-objective quality-diversity algorithms can help explore such vast spaces. We adopt the Dominated Novelty Search (DNS) algorithm \cite{bahlous2025dominated}, which reformulates quality-diversity optimization as a genetic algorithm with local competition implemented through fitness transformations rather than explicit archive structures. Extending this framework to a multi-objective setting, we introduce Pareto Dominated Novelty Search, which replaces single-objective fitness comparisons with Pareto dominance relations.

This algorithm maintains a population of N candidate designs $\mathbf{X} = (\mathbf{x}_i)_{i=1}^{N}$, where each design $\mathbf{x}_i$ represents a gripper encoded in the graph design space described in \cref{sec:space}. Each design is characterized by:

\begin{itemize}
\item A vector of objective values $\mathbf{f}_i \in \mathbb{R}^m$ measuring performance across \textit{m} competing objectives
\item A behavioural descriptor $\mathbf{d}_i \in \mathbb{R}^D$ capturing meaningful features of the solution's behaviour
\end{itemize}

Similarly to the DNS algorithm, we follow the workflow illustrated in \cref{fig:fig2}.A: At each generation, the population undergoes reproduction (selection, crossover, and mutation), offspring are evaluated to obtain objective values $\mathbf{f}_i$ and behavioural descriptors $\mathbf{d}_i$, competition fitness scores $\tilde{f}_i$ are computed for the augmented population, and individuals are sorted and truncated to maintain population size $N$.

\subsection{Competition fitness}

We extend the original definition of competition fitness in the DNS algorithm by employing Pareto dominance instead of single-objective direct fitness comparisons. The resulting competition fitness condenses diversity and multiple objectives into a single scalar metric and is computed as follows:

\begin{enumerate}
    \item For each solution \textit{i}, construct the set of all solutions that Pareto-dominate it (\cref{fig:fig2}.B):
\begin{equation}
\mathcal{D}_i = \{ j \in \{1, \ldots, N\} \mid \mathbf{x}_i \succ \mathbf{x}_j \}
\end{equation}
    where $\mathbf{x}_i \succ \mathbf{x}_j$ refers to Pareto dominance, which states that solution $\mathbf{x}_j$ dominates solution $\mathbf{x}_i$ if and only if:
    \begin{equation}
    \forall k \in \{1,\ldots,m\}: f_i^{(k)} \geq f_j^{(k)}  \text{ and } \exists k: f_i^{(k)} > f_j^{(k)}
    \end{equation}
    \item Calculate pairwise distances in descriptor space (\cref{fig:fig2}.C) between solution \textit{i} and all dominating solutions:
\begin{equation}
\delta_{ij} = \|\mathbf{d}_i - \mathbf{d}_j\| \quad \forall j \in \mathcal{D}_i
\end{equation}
    \item The competition fitness is defined as the average distance to the \textit{k}-nearest dominating solutions:
\begin{equation}
\tilde{f}_i = \begin{cases}
\frac{1}{k} \sum_{j \in \mathcal{K}_i} \delta_{ij} & \text{if } |\mathcal{D}_i| > 0 \\
+\infty & \text{otherwise}
\end{cases}
\end{equation}
where $\mathcal{K}_i \subseteq \mathcal{D}_i$ contains the indices of the \textit{k} dominating solutions with smallest distances to solution \textit{i}. Solutions on the Pareto front ($\mathcal{D}_i = \emptyset$) receive infinite competition fitness, ensuring their preservation.
\end{enumerate}

\subsection{Genetic operators}
The evolutionary search follows the process in \cref{fig:fig2}.A, and employs a combination of selection, crossover, and mutation operators designed specifically for our graph-based design representation.

\textbf{Selection.} We employ tournament selection of size $\kappa$ operating in fitness space. For each parent selection, $\kappa$ individuals are randomly sampled and compared using 1-step Pareto dominance: if any individual is non-dominated, it is selected; otherwise, one is chosen at random.

\textbf{Crossover.} Offspring are generated using node-wise two-point crossover (CX2Point).

\textbf{Mutation.} We apply a diverse set of mutation operators operating at parametric and structural levels. Parametric mutations modify continuous variables by adding Gaussian noise to either the position $\boldsymbol{x}_i$, edge width $w_{th}$, or distance multipliers $w_d$ of individual nodes. Structural mutations modify the graph topology:
\begin{itemize}
\item Add-node mutation: Inserts a new node with random initialization
\item Remove-node mutation: Deletes a randomly selected node
\item Split-node mutation: A node is split in two which are placed diametrically opposed relative to the original
\end{itemize}

\textbf{Culling.} Following evaluation and competition fitness evaluation, individuals are ranked by $\tilde{f}_i$ and the top $\tilde{n}$ are retained for the next generation.

\subsection{Behavioural descriptor}

To guide diversity-based optimization, we extract a modular set of behavioural descriptors capturing geometric, topological, and parametric aspects of the graph-based design. The descriptor comprises several feature groups: 
(1) node-spatial features (e.g. centroid, spread), 
(2) node-attribute features (e.g. thickness and distance weight statistics), 
(3) graph features (e.g. edge density, relative size, assortativity, clustering metrics, degree entropy, spectral properties), and 
(4) geometry features (e.g. volume fraction, compactness, geometric centroid).  

Most graph-based features are naturally bounded—for instance, edge density $[0,1]$, assortativity $[-1,1]$, and normalized entropy $[0,log_2(N)]$ are all bound, facilitating consistent distance metrics in descriptor space. We note that these descriptors characterize phenotypic diversity and do not explicitly account for functional diversity, which is inherently problem-specific.

\section{Fitness Evaluation}
\label{eval}

We employ a finite strain Neo-Hookean hyperelasticity formulation to capture the non-linear material response of soft elastomers under large deformations. Contact interactions use the Incremental Potential Contact (IPC) method \cite{Li2020IPC}, which guarantees intersection-free configurations throughout simulation, which is essential for evolutionary optimization where designs may exhibit pathological geometries. Temporal discretization employs an implicit Newmark scheme well-suited to the near-quasi-static deformation regime of robotic grasping. The resulting non-linear systems are solved using Newton-Raphson iteration with SparseLU factorization for the linearized sub-problems. Spatial discretization uses first-order tetrahedral elements with typical mesh resolutions of $5\cdot10^3$ elements per design.

For each simulation, we extract contact forces $f_c$, contact area $A_c$, object strain energy $\mathcal{U}$, and end-effector displacements $\boldsymbol{u}$ as the basis for fitness evaluation. A key advantage of this physics-based formulation is direct optimization of task-relevant metrics (contact forces, compliant deformation) rather than requiring designer expertise to translate manipulation requirements into equivalent load systems \cite{pinskier2024diversity, liu2024topology}.

\section{Results}

The results presented in this section make use of a series of powerful and convenient open-source software packages: \href{https://shapely.readthedocs.io}{Shapely} and \href{https://cadquery.readthedocs.io/en/latest/}{CadQuery} for 2D and 3D geometric manipulation, fTetWild \cite{ftetwild} for meshing, DEAP \cite{deap} for prototyping evolutionary algorithms and PolyFEM \cite{polyfem} for finite element modelling.  

\begin{figure}[h]
    \centering
    \vspace{-0.2cm}
    \includegraphics[width=\linewidth]{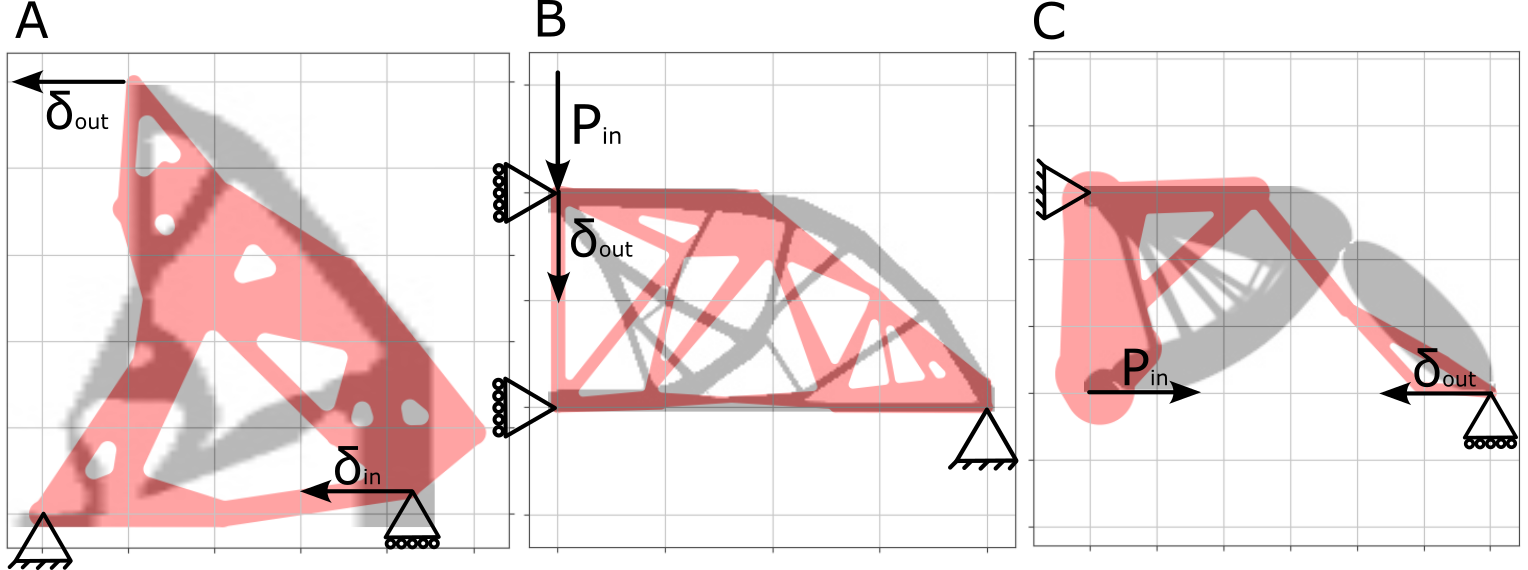}
    \caption{A. Soft gripper problem $min\{\delta_{out} \, | \, V^* = 0.3\}$, B. Inverter problem $min\{-  \delta_{out} \, | \, V^* = 0.2\}$, C. MBB beam problem  $min\{-\delta_{out} \, | \, V^* = 0.3\}$. }
    \vspace{-12pt}
    \label{fig:res_topo}
\end{figure} 

\subsection{Graph-space expressiveness}

Before presenting multi-objective soft gripper optimization results, we first validate that a graph-based representation can capture diverse optimal structures. We apply this representation to three standard structural optimization benchmarks \cite{sigmund1997design} using linear FEM as the function evaluator.

\Cref{fig:res_topo} presents optimal solutions obtained for: (A) a soft gripper problem minimizing output displacement subject to $V^* = 0.3$ volume fraction constraint, (B) an centrally loaded beam problem minimizing compliance with $V^* = 0.3$, and (C) an inverter mechanism maximizing output displacement with $V^* = 0.2$. In each case, our approach successfully recovers solutions that approximate known optimal topologies from traditional topology optimization methods (shown as shaded backgrounds in \cref{fig:res_topo}).

These results demonstrate that the graph edge inference mechanism provides sufficient expressiveness to capture diverse optimal structures across different problem classes. The representation's ability to approximate both compliance-based and stiffness based problems validates its suitability for the more complex multi-objective soft gripper design task that follows.

\subsection{Grasping cases}
\label{sec:cases}

\begin{figure}[b]
    \centering
    \includegraphics[width=0.8\columnwidth]{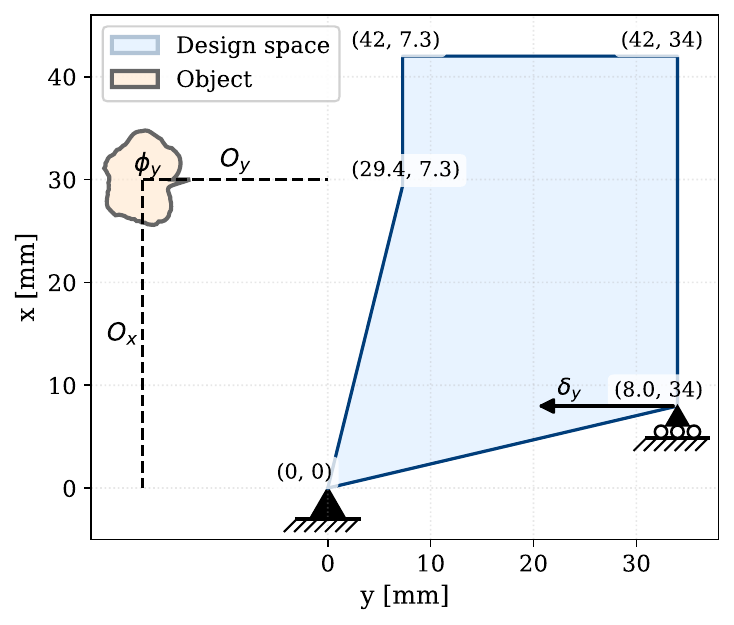}
    \vspace{-12pt}
    \caption{Generic grasping problem description, dimensions of the optimisation domain space, boundary conditions and location of target objects.  }
    \vspace{-0pt}
    \label{fig:cases}
\end{figure}

We evaluate the proposed framework on planar grasping tasks derived from the Aloha parallel gripper geometry \cite{Aloha}. The design domain (\cref{fig:cases}) defines the admissible region for node placement, with Dirichlet boundary conditions applied such that each finger operates as a parallel compliant mechanism. Grasp closure is actuated by prescribed vertical displacement $\delta_y$ at the sliding boundary, with no explicit control over contact location. Pull strength is evaluated by applying horizontal displacement $\delta_x$ while the object remains fixed.

We define four grasping scenarios spanning two grasp types (pinch and power) and two object properties (rigid and compliant). Pinch gripps involve small objects contacted near the fingertip; power gripps engage larger objects at the finger mid-section. Compliant objects have Young's modulus one-third that of the gripper material. All objects are spherical with geometric parameters given in \cref{tab:cases}.

For each design $\mathbf{x}_i$ and scenario $s \in \{1,2,3,4\}$, displacement-controlled FEM simulation yields time histories of contact force $f_c(t)$, contact area $A_c(t)$, and object strain energy $\mathcal{U}(t)$. Each scenario defines an objective function measuring gripp quality:

Pinch-rigid (small rigid object) maximizes pull force:
\begin{equation}
f^{(4)}(\mathbf{x}_i) = \int_{T_{\mathrm{pull}}} f_c(t)\,\mathrm{d}t
\end{equation}

Power-rigid (large rigid object) maximizes contact area to favour surface conformance:
\begin{equation}
f^{(1)}(\mathbf{x}_i) = \int_{T_{\mathrm{pull}}} A_c(t)\,\mathrm{d}t
\end{equation}

Compliant objects balance strain energy minimization (avoiding excessive deformation of the object) with pull force:
\begin{equation}
f^{(2,3)}(\mathbf{x}_i) = \lambda_{\mathcal{U}} \cdot \max(\mathcal{E}(t)) + \int_{T_{\mathrm{pull}}} f_c(t)\,\mathrm{d}t
\end{equation}
where $\lambda_{\mathcal{U}}=500$ normalizes the terms based on preliminary simulations.

The complete multi-objective problem minimizes:
\begin{equation}
\mathbf{f}(\mathbf{x}_i) = \left[ f^{(1)}(\mathbf{x}_i), f^{(2)}(\mathbf{x}_i), f^{(3)}(\mathbf{x}_i), f^{(4)}(\mathbf{x}_i) \right]^T \in \mathbb{R}^4
\end{equation}
subject to $V^* \leq V_{max}/V_{total}$, where $V^*(\mathbf{x}_i)$ is the volume fraction, $V_{total}$ is the domain area, and $V_{max}$ is the imposed volume limit.

\begin{table}[]
\centering
\caption{Complimentary table to \cref{fig:cases} for different grasping problems. }
\label{tab:cases}
\begin{tabular}{lcccc}
\hline
&
\shortstack{\textbf{Large rigid} \\ $f^1$} &
\shortstack{\textbf{Large soft} \\ $f^2$} &
\shortstack{\textbf{Small soft} \\ $f^3$} &
\shortstack{\textbf{Small rigid} \\ $f^4$} \\
\hline
$O_y$ (mm) & -18 & -18 & -18 & -18 \\
$O_{x}$ (mm) & 28 & 28 & 37 & 37 \\
$\phi$ (mm) & 35 & 35 & 10 & 1.5 \\
$\delta_x$ (mm) & 10 & 10 & 16 & 18 \\
\hline
\vspace{-22pt}
\end{tabular}
\end{table}

\subsection{Pareto-optimal solutions}
\label{sec:pareto}

\begin{figure}[]
    \centering
    \includegraphics[width=1\columnwidth]{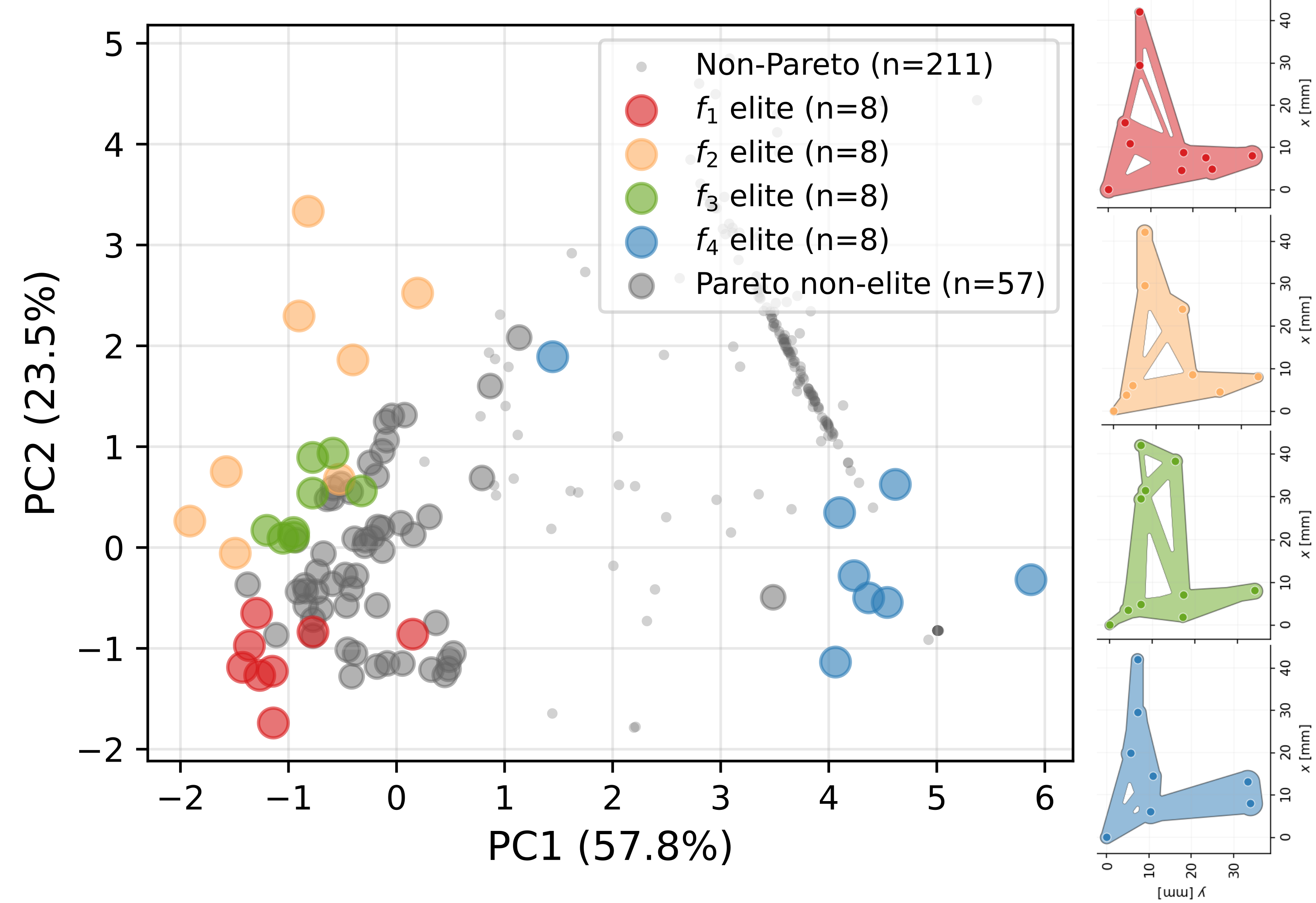}
    \vspace{-22pt}
    \caption{Pareto front projected onto the first two principal components of objective space. Colour highlighting elite solutions (top-performing individuals for each grasping scenario) reveals clustering associated with objective specialisation. Representative elite individuals colour-coded according to specialisation.}
    \vspace{-12pt}
    \label{fig:pareto}
\end{figure}

We optimized the multi-objective problem from \cref{sec:cases} using a population of $N=300$ individuals over 200 generations with a 10-dimensional behavioural descriptor and fixed computational time for fitness evaluation. The resulting Pareto front contains 89 non-dominated solutions, projected onto the first two principal components (capturing 81.3\% variance) in \cref{fig:pareto}. 

Clear clustering emerges by objective specialization despite no explicit fitness-space declustering mechanism, indicating that phenotype-level diversity preservation effectively translates into functional diversity. The projection reveals a primary trade-off: PC1 (57.8\% variance) separates the small rigid object scenario ($f^4$) from others, reflecting that soft grippers which typically rely on large contact areas struggle to generate high forces on small rigid objects. 

Elite individuals (top-8 performers per objective) exhibit distinct morphological strategies. All solutions share a common compliant mechanism: the lower beam (linking the two Dirichlet nodes) bends upwards, driving vertically-oriented beams that connect to the contact region. However, designs differ in moment arm length and contact geometry. Large-object specialists ($f^1$, $f^2$) feature longer moment arms, prioritizing contact force over tip displacement. Small-object specialists ($f^3$, $f^4$) use shorter moment arms for greater tip displacement. Furthermore, $f^3$ specialized designs develop hook-like structures that engage deformable objects without crushing, while $f^1$ specialized designs favour thin, compliant engagement elements for maximum compliance and contact area.

\begin{figure*}[]
    \centering
    \includegraphics[width=1\textwidth]{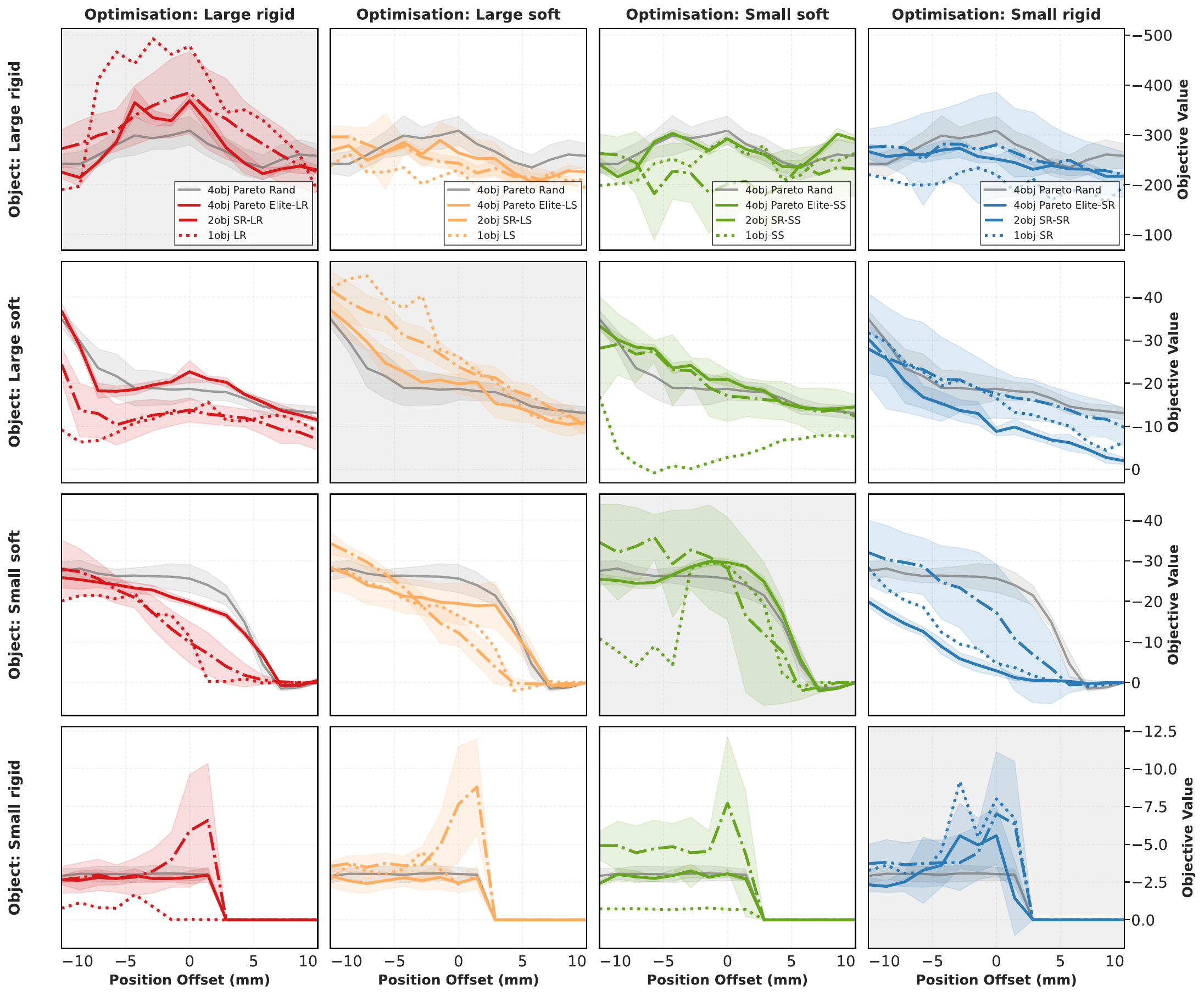}
    \caption{Robustness evaluation across grasping scenarios and object position perturbations. Each column represents designs optimized for a specific objective or selected for specialising in said objective. Each row tests designs on a specific scenario. Diagonal plots (grey shaded) show specialists on their training scenario; off-diagonal plots show performance against other scenarios. Curve types: solid lines show elite specialists from 4-objective optimization, dashed lines show 2-objective co-optimization including the column's objective, dotted lines show single-objective optimization, and grey curves (repeated in all plots) show the random set of 4-objective Pareto designs. Shaded regions indicate standard deviation in fitness ($\pm\sigma$). More negative objective values indicate better performance.}
    \label{fig:grid}
    \vspace{-14pt}
\end{figure*}

Correlation analysis provides further insight: population-wide correlations are moderate ($r\approx0.7$), but Pareto-front correlations weaken and several reverse sign ($r_{i,4}\approx-0.5$). This indicates that while objectives co-improve during design space exploration, they impose conflicting requirements at the performance frontier. Consequently, the four scenarios define distinct, irreducible performance dimensions that prevent convergence to a single optimal design. 

\subsection{Robustness to perturbations}

To investigate whether diversity-driven multi-objective optimization produces designs that generalize beyond their training scenarios, we conduct a robustness analysis comparing designs optimized under different objective combinations. We performed eight optimization runs: one 4-objective run (as in \cref{sec:pareto}), four 1-objective runs (one per scenario), and three 2-objective runs pairing $f^4$ with each other objective (since $f^4$ is previously shown to conflict with all other scenarios). 

From these runs, we extracted 13 design sets for evaluation: (1) a random sample of 30 designs from the 4-objective Pareto front; (2) elite specialists from the 4-objective run (8 top performers per objective); (3) random samples of 20 designs from each 2-objective Pareto front; and (4) the single best design from each 1-objective run. Each design set was evaluated on all four grasping scenarios under positional perturbations: object positions varied by $\pm10$mm in $O_x$ from nominal values (\cref{tab:cases}), where negative perturbations move objects toward power gripp positions (closer to the finger base), while positive perturbations move toward pinch positions (closer to the fingertip).

Results are shown in \cref{fig:grid}, where columns correspond to the optimization objective(s) and rows to the test scenario. Diagonal plots (shaded) show specialists/optima tested on their training scenario; off-diagonal plots reveal generalization to either unseen or off-speciality scenarios. The grey curves show the random 4-objective Pareto sample as a baseline across all tests.

The robustness analysis reveals three key findings supporting diversity-driven generalization. First, specialization comes at the cost of brittleness: single-objective optima dominate their training scenarios (diagonal plots) but exhibit sharp performance drops when tested on other scenarios. This overfitting to specific object positions is visible only in 1-objective designs, confirming that greedy optimization produces brittle solutions.

Second, multi-objective optimization mitigates overfitting: 2-objective designs show no positional overfitting, and 4-objective designs maintain competitive performance across all scenarios. While specialists naturally dominate their training cases (diagonal: 1-obj $>$ 2-obj $>$ 4-obj), the performance gap narrows substantially off-diagonal.

Finally, diversity appears to support transferable grasping behaviour. The random 4-objective Pareto sample (grey) achieves strong off-diagonal performance and outperforms all specialist sets on large-rigid or small soft tests. These results suggest that the diverse Pareto front contains designs that exploit adaptable structural strategies rather than relying on scenario-specific tuning. Together, these observations indicate that optimization across a finite and diverse set of grasping scenarios can promote robustness to novel object properties and contact conditions.

\begin{figure}[]
    \centering
    \includegraphics[width=1\columnwidth]{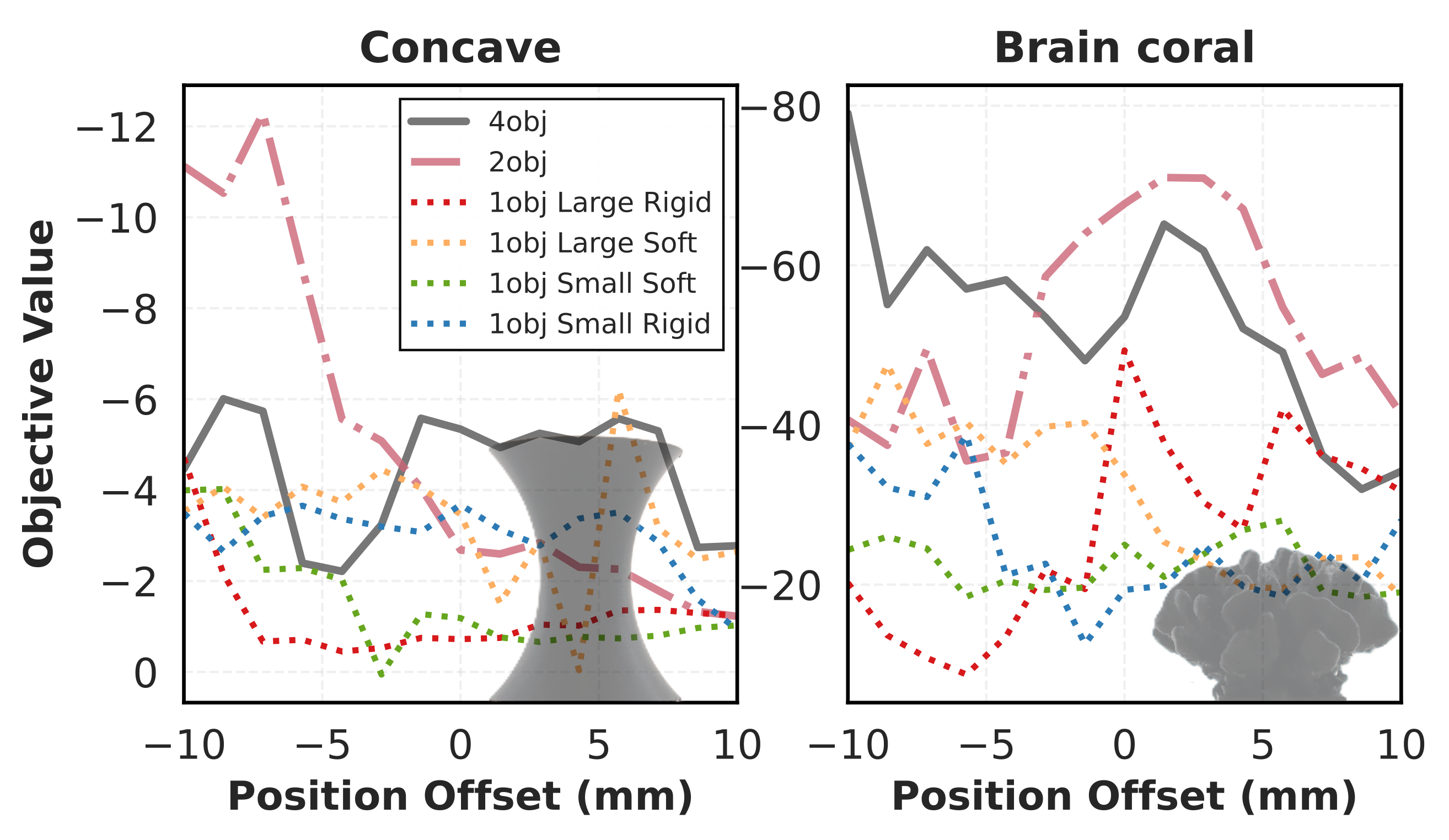}
    \vspace{-12pt}
    \caption{Best-performing individuals from single-, two-, and four-objective optimization runs evaluated on two novel object geometries shown in the background.}
    \vspace{-16pt}
    \label{fig:xtra}
\end{figure}

To further probe generalization beyond the training scenarios, we evaluate the same design sets on two previously unseen objects: a concave column and a coral-like structure (\cref{fig:xtra}). Rather than comparing entire populations, we identify the best-performing individual within each set—namely, the 4-objective run, the three 2-objective runs, and the four single-objective runs. Performance is assessed using a position-invariant score defined as the mean objective value across all positional perturbations.

As shown in \cref{fig:xtra}, designs optimized for a single objective exhibit limited transfer, with performance degrading substantially on both novel objects. In contrast, both 2-objective and 4-objective runs contain individuals that achieve performance comparable to that observed on the original grasping scenarios.

Within the 2-objective runs, the top-performing individuals for both novel objects consistently originate from the large-rigid optimization set, suggesting that these objects impose mechanical requirements similar to the large-rigid scenario. However, in the 4-objective run, the best-performing individuals are drawn from the random Pareto subset rather than from specialist elites. This indicates that maintaining population-level diversity enables the discovery of designs whose grasping strategies transfer more effectively to previously unseen object geometries.

\subsection{Physical validation}

A limited experimental validation is executed to ground the behaviour of the obtained optimal grippers in reality. Two designs selected from the Pareto front of the four-objective run were fabricated in Formlabs Elastic 50A and mounted on a parallel gripper driven by a stepper motor (XM430-W210, DYNAMIXEL, South Korea). The gripper assembly was attached to a linearly actuated platform to control vertical displacement during pull-out tests. A load cell (H3-C3-25KG-3B, Zemic, Netherlands) recorded extraction forces. grasping performance was qualitatively evaluated on objects of varying geometry: a sphere, a coral-like structure, and a pin.

\begin{figure}[]
    \centering
    \includegraphics[width=1\columnwidth]{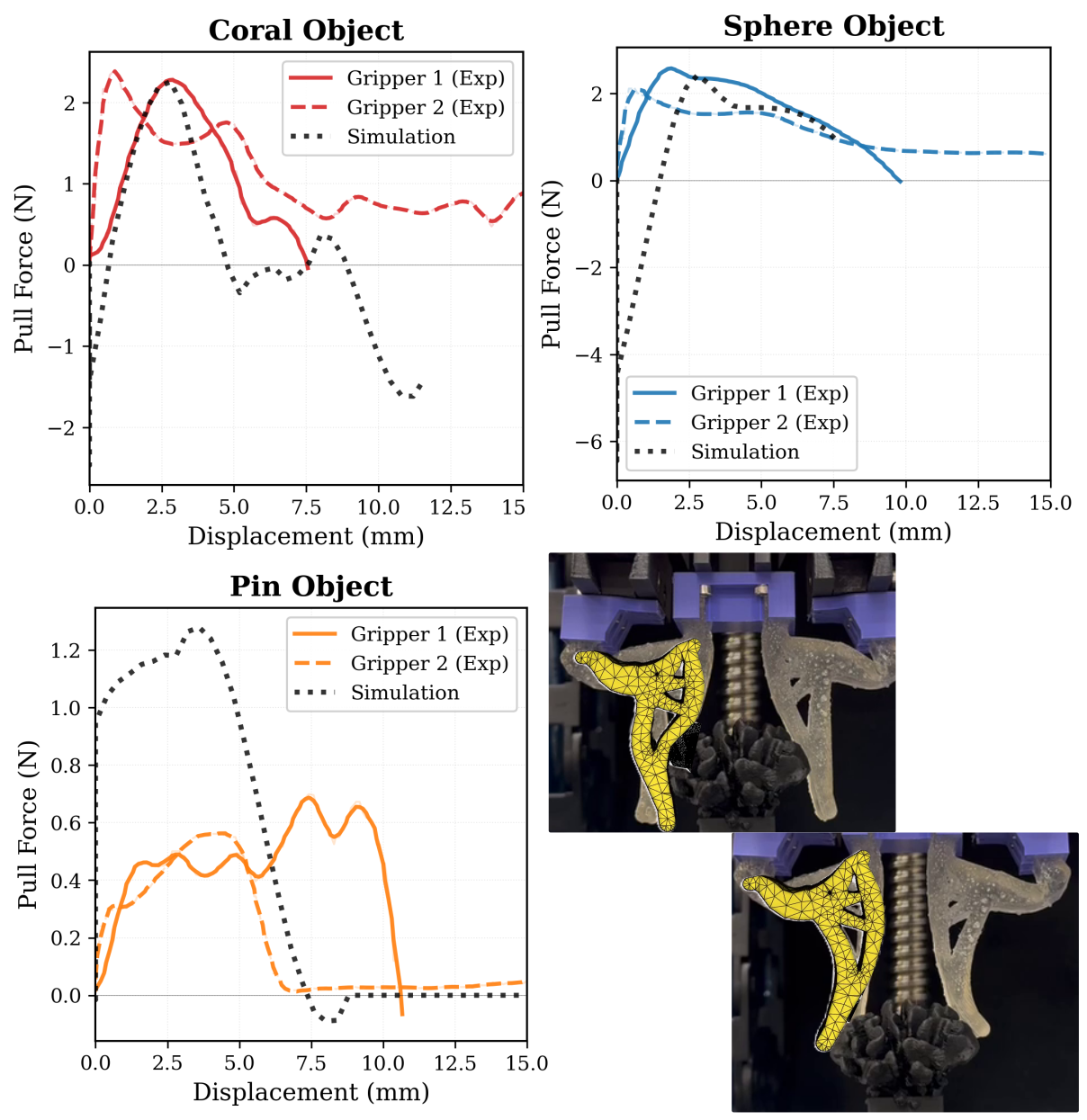}
    \vspace{0pt}
    \caption{Direct comparison of pull forces in several grasping scenarios, and still images comparing the simulation and experimental run of a coral grasping task. }
    \vspace{-16pt}
    \label{fig:exp}
\end{figure}

While these experiments are limited in scope, they verify that the optimized designs exploit deformation and contact mechanisms predicted by simulation, and that the observed grasping trends, including approximate pull-force magnitudes, transfer qualitatively to physical hardware (\cref{fig:exp}).

\section{Conclusion}

This work introduces a framework for near-freeform design of planar soft grippers that combines a graph-based design representation, high-fidelity finite-element simulation, and a multi-objective quality-diversity optimisation strategy. In particular, we extend Dominated Novelty Search to multi-objective settings, enabling exploration of expressive design spaces across multiple grasping scenarios while remaining compatible with physics-based evaluation.

We show that the proposed graph representation can recover diverse, near-optimal topologies and encode a wide range of compliant gripper morphologies. Multi-objective diversity-driven optimisation produces Pareto fronts containing both specialised and robust designs, with phenotype-level diversity translating into functional diversity in objective space. Robustness analyses further indicate that optimisation across multiple grasping scenarios mitigates overfitting observed in single-objective designs and promotes transfer to off-training conditions and previously unseen object geometries.

These findings suggest that diversity-driven optimisation provides a principled pathway toward more general-purpose soft manipulation, allowing robust mechanical strategies to emerge from a limited set of objectives rather than scenario-specific tuning. A natural extension of this work would focus on larger-scale experimental validation and integration with learning-based control to evaluate how mechanically generalised grippers interact with perception and control policies.

Let $W_{ij} = \big(\lVert \mathbf{x}_i - \mathbf{x}_j\rVert / L + w_{ij}\big)/d_{\max}$, and call edge $(i,j)$ active when $W_{ij} < 1.$ Write $E(G) = \{(i,j) : W_{ij} < 1\}$ for the active edge set.

\bibliographystyle{plainnat}
\bibliography{refs}

\end{document}